\documentclass[letterpaper, 10 pt, journal, twoside]{IEEEtran}

\IEEEoverridecommandlockouts                              

\usepackage{graphics} 
\usepackage[caption=false,font=footnotesize]{subfig}
\usepackage{epsfig} 
\usepackage{gensymb}
\usepackage{cite}
\usepackage{amsmath}
\usepackage{amssymb}
\usepackage{amsfonts}
\usepackage{algorithm}
\usepackage{algorithmic}
\usepackage{tensor}
\usepackage{textcomp,booktabs}
\usepackage[usenames,dvipsnames]{color}
\usepackage{makecell}
\usepackage{colortbl}
\usepackage{xcolor}
\usepackage{multirow}
\usepackage{color,soul}
\usepackage{float}
\definecolor{mygray}{gray}{.9}
\usepackage{csquotes}

\usepackage[flushleft]{threeparttable}

\title{Learning Deep Nets for Gravitational Dynamics with Unknown Disturbance through Physical Knowledge Distillation: Initial Feasibility Study
}
\author{Hongbin Lin$^{1}$, Qian Gao$^{2}$, Xiangyu Chu$^{1}$, Qi Dou$^{3}$, Anton Deguet$^{4}$, Peter Kazanzides$^{4}$, and K. W. Samuel Au$^{1}$
\thanks{Manuscript received: October, 13, 2020; Revised January, 13, 2021; Accepted February, 8, 2021.}

\thanks{This paper was recommended for publication by Editor Pietro Valdastri upon evaluation of the Associate Editor and Reviewers' comments. This work was supported in part by the CUHK Chow Yuk Ho Technology Centre of Innovative Medicine, in part by the Multi-Scale Medical Robotics Centre, InnoHk, 8312051, RGC T42-409/18-R, in part by SHIAE (BME-p1-17), and in part by the Natural Science Foundation of China under Grant U1613202. (Corresponding author: K. W. Samuel Au)}
\thanks{$^{1}$Hongbin Lin, Xiangyu Chu and K. W. Samuel Au are with Department of Mechanical and Automation Engineering, The Chinese University of Hong Kong, Hong Kong. {\tt\small hongbinlin@link.cuhk.edu.hk; xychu@mae.cuhk.edu.hk; samuelau@cuhk.edu.hk}}%
\thanks{$^{2}$Qian Gao is with Shenzhen Institute of Artificial Intelligence and Robotics for Society, School of Science and Engineering, The Chinese University of Hong Kong, Shenzhen, China.{\tt\small qiangao@cuhk.edu.cn}} %
\thanks{$^{3}$Qi Dou is with Department of Computer Science and Engineering, The Chinese University of Hong Kong, Hong Kong.
{\tt\small qdou@cse.cuhk.edu.hk}} %
\thanks{$^{4}$Anton Deguet and Peter Kazanzides are with Department of Computer Science, Johns Hopkins University, USA.
{\tt\small anton.deguet@jhu.edu; pkaz@jhu.edu}} %

\thanks{Digital Object Identifier (DOI): see top of this page.} } 

\begin{document}

\maketitle

\begin{abstract}

Learning high-performance deep neural networks for dynamic modeling of high Degree-Of-Freedom (DOF) robots remains challenging due to the sampling complexity. Typical unknown system disturbance caused by unmodeled dynamics (such as internal compliance, cables) further exacerbates the problem. In this paper, a novel framework characterized by both high data efficiency and disturbance-adapting capability is proposed to address the problem of modeling gravitational dynamics using deep nets in feedforward gravity compensation control for high-DOF master manipulators with unknown disturbance. In particular, Feedforward Deep Neural Networks (FDNNs) are learned from both prior knowledge of an existing analytical model and observation of the robot system by Knowledge Distillation (KD). Through extensive experiments in high-DOF master manipulators with significant disturbance, we show that our method surpasses a standard Learning-from-Scratch (LfS) approach in terms of data efficiency and disturbance adaptation. Our initial feasibility study has demonstrated the potential of outperforming the analytical teacher model as the training data increases.

\end{abstract}

\section{INTRODUCTION}

\IEEEPARstart{M}{odel-based} control for modern robotics, such as compliance control, impedance control and hybrid force control, endows robots with the ability to deal with interaction with the environment. The performance of such control strategies highly depends on the accuracy of dynamic models of robots, which are capable of reasoning about joint acceleration and joint torques (forward and inverse dynamics).

\begin{figure}[!tbp]
\centering
\includegraphics[width=3.4in]{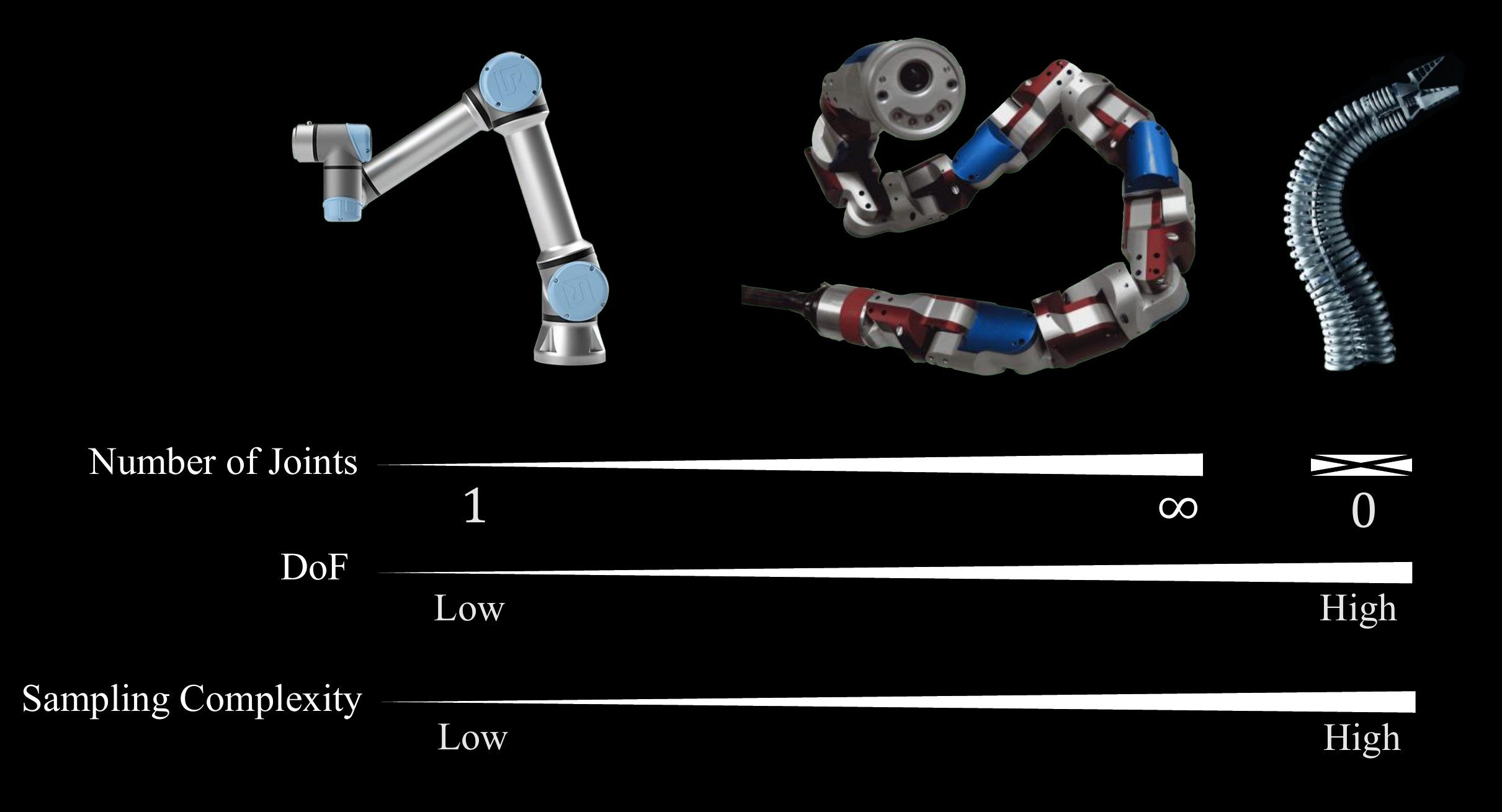}
\caption{Comparing the number of joints, Degrees of Freedom (DOF) and sampling complexity among traditional rigid-link robots, serpentine robots and soft robots/continuum manipulators. Robots in the figure from left to right are a UR5 robot arm, a CMU snake robot\cite{wright2012design} and a Festo manipulator\cite{falkenhahn2015model}, representing rigid-link robots, serpentine robots and soft robots/continuum manipulators, respectively.}
\label{fig:robot_overview}
\end{figure}

Conventionally, dynamic models can be derived by physics law analytically\cite{siciliano2009robotics}. Such physics-based analytical models, parameterized by kinematic and dynamic properties of robot system (such as mass, length of links, the center of mass, \textit{etc.}), are able to describe a global mapping over the entire state space of the robot system. The parameters of the analytical models can be learned by offline approaches (\textit{e.g.}, system identification) and online approaches (\textit{e.g.}, adaptive control)\cite{wu2010overview}. However, the analytical solutions cannot easily adapt to high system disturbance caused by unmodeled dynamics (such as internal compliance, cables, and so on) due to their
fixed basis function\cite{nguyen2010using}. In such a case, extensive engineering efforts are required to further model the disturbance for specific systems, which is an arduous process for a robot system with complex disturbance \cite{Klu2018case}.

To overcome the limitation of the physics-based approaches, researchers start to turn their attention to model-free approaches leveraging machine learning techniques as alternative solutions. Particularly, Linear Regression\cite{haruno2001mosaic}\cite{schaal2002scalable}, Gaussian Mixture Regression\cite{khansari2011learning}, Gaussian Process Regression\cite{kocijan2004gaussian}, Support Vector Machine\cite{choi2007local}, Feedforward Neural Network\cite{jansen1994learning,yu2000pd,yilmaz2020neural} and Recurrent Neural Network\cite{polydoros2015real}\cite{rueckert2017learning} were applied to approximate the system dynamics of robots based on measured data. These methods, known as Learning from Scratch (LfS),  did not require any prior physics knowledge. Furthermore, LfS was potentially capable of capturing the unknown nonlinearity of system disturbance due to its higher model capacity compared to the physics-based approach. However, achieving high performance over the entire state space of the robot system using the LfS approach for high Degree-Of-Freedom (DOF) robots still remained challenging\cite{nguyen2011model}. Sampling data using the LfS approach was required to cover the majority of the state space with adequate density. This requirement was hard to reach for high-DOF systems, since the sampling complexity, $O(N^n)$, exponentially increased with the dimensionality of the state space, where $n$ is the number of DOF and $N$ is sampling points for each DOF. Fig. \ref{fig:robot_overview} shows the sampling complexity for rigid-link robots, serpentine robots\cite{wright2012design} and soft robots\cite{della2020model}/continuum manipulators\cite{falkenhahn2015model}, illustrating the challenge of sampling complexity for high-DOF robots. In practice, the sampling problem prevents practitioners from applying the LfS approach to high-DOF robots considering the expense of data generation (\textit{e.g.}, mechanical maintenance, long operation time).

To enjoy merits from the physics-based and the LfS approaches, researchers attempted to explore hybrid approaches by combining these methods. One thread was error-learning frameworks where the error of the physics-based model was learned by Locally Weighted Projection Regression\cite{petkos2006learning}\cite{de2012line}, Gaussian Process Regression \cite{nguyen2011model,camoriano2016incremental} and Feedforward Neural Network\cite{kappler2017new}. Although the error-learning frameworks enhanced the model capacity compared to the pure physics-based approach for adapting to a high-disturbance system, they suffered from low data efficiency similar to that of the standard LfS approach since the errors over the entire state space of a system were required to be learned from scratch without any prior. Therefore, they did not solve the problem of sampling complexity fundamentally due to their LfS learning fashion. 

Another thread of the hybrid approach was learning system dynamics with estimators encoded by prior knowledge of physics inference. Dynamic knowledge (\textit{e.g.}, Newton-Euler inference\cite{ledezma2017first} and Euler-Lagrange inference\cite{lutter2019deep}) was encoded into the topology of neural networks. These approaches showed higher data efficiency compared to the standard LfS approach. However, the ability to adapt to the unknown disturbances remained poor since the model capacity of physics-encoded approximators was drastically bounded by the pre-encoded physics prior. As the result, further engineering efforts of modeling disturbance were required to minimize the model bias caused by unmodeled dynamics, which can hardly alleviate the engineering burden compared to those of the physics-based approach.

To disentangle the dynamic modeling problem for high-DOF robot systems with unknown disturbance, we propose a data-efficient, disturbance-adapting hybrid approach using a knowledge fusion technique, Knowledge Distillation (KD)\cite{hinton2015distilling}. KD was originally proposed to solve the problem of model compression\cite{bucilua2006model}, where the knowledge of a large cumbersome model or ensemble of models, called Teacher Model (TM), is transferred to a small efficient model, called Student Model. For KD, Student Model learned from TM by optimizing a weighted average of two objective functions, the loss function of Student Model with TM and that with training data.
It is widely used in AI community due to its high efficiency in inheriting the knowledge from the large cumbersome model (\textit{e.g.} the generalization) as well as learning from new data. In this paper, the student Feedforward Deep Neural Network (FDNN) is learned from the knowledge of an existing physics-based TM, called Physical Teacher Model (PTM), and the measured data for learning one of the classical components of dynamics, gravitational dynamics, in high-DOF rigid-link robots. The learned model is further applied to feedforward Gravity Compensation Control (GCC) for evaluating its performance.

To our best knowledge, we are the first to demonstrate the efficacy of KD in the field of dynamics modeling for 6-DOF rigid-body manipulators. Moreover, it can be potentially applied to the robot systems with higher DOF (such as soft robots/continuum manipulators), which is our future research direction. We have developed a fully-automatic open-source software package for the reproduction of our method and experiment\cite{ExperimentCode}. 
We tested our method on 3 robot manipulators to verify the algorithm's reliability or adaptability to handle the variations in robots.

Overall, our contributions are:

\begin{enumerate}
    \item Proposal of a novel knowledge-fusion learning framework for modeling gravitational dynamics of a robot system using FDNN, which pre-processes input-output signals of FDNN by both trigonometric representation and normalization, and learns from PTM and observed data using KD. Our method is applicable for GCC in high-DOF robot systems with unknown highly nonlinear configuration-dependent and direction-dependent disturbance given sparse training data.

    \item Validation of the effectiveness of our method in master manipulators with significant configuration-dependent and direction-dependent disturbance. Our method exhibits higher data efficiency and disturbance-adapting capability compared to the standard LfS method as well as the potential to outperform the PTM with the increase of training data. 
\end{enumerate}

\section{Background}
Dynamics for rigid-link robots can be formulated as
\begin{equation} 
\label{equ:rgd}
    \boldsymbol{\tau} = \boldsymbol{M}(\boldsymbol{q})\boldsymbol{\ddot{q}} + \boldsymbol{C}(\boldsymbol{q},\boldsymbol{\dot{q}})\boldsymbol{\dot{q}} + \boldsymbol{f_f}(\boldsymbol{\dot{q}})+\boldsymbol{g}(\boldsymbol{q})+ \boldsymbol{J}(\boldsymbol{q})^{T}\boldsymbol{h_e} + \boldsymbol{\epsilon},
\end{equation}
where $\boldsymbol{q}$,$\boldsymbol{\dot{q}}$,$\boldsymbol{\ddot{q}}$ denote joint position, velocity and acceleration, respectively; $\boldsymbol{M}$ and $\boldsymbol{C}$ are a inertia matrix and a matrix of the centrifugal and Coriolis force, respectively; $\boldsymbol{f_f}$
is a force vector collecting the Coulomb and viscous friction, $\boldsymbol{g}$ represents gravity, $\boldsymbol{\epsilon}$ represents the disturbance due to unmodeled dynamics, $\boldsymbol{J}$ is a Jacobian matrix and $\boldsymbol{h_e}$ is a vector representing forces, moments exerted by the environment on the end-effector.

To compensate the dynamic effect of gravity in a robot system, a feedforward model-based GCC can be applied given the dynamics model of gravity. The required compensated torques $\boldsymbol{\tau_c}$ in GCC is given as
\begin{equation}
\label{equ:gcc_only_gravity}
    \boldsymbol{\tau_c} = \boldsymbol{g}(\boldsymbol{q}).
\end{equation}

 To estimate the gravity in (\ref{equ:gcc_only_gravity}), one can utilize a physics-based, analytical model derived by either the Euler-Lagrange approach or Newton-Euler approach\cite{siciliano2009robotics}. Parameters for the physics-based model can be obtained by standard Least Square Estimation \cite{ma1996identifying} based on the linearity of the analytical model\cite{siciliano2009robotics}, if $\boldsymbol{\epsilon}$ is negligible.
 
However, when the disturbance in the robot system is significant, the performance using the physics-based model is relatively poor due to its weak disturbance-adapting capability. In our previous study, significant joint drifts of a master manipulator, indicating poor control performance, were observed in a Drift Test when evaluating the naive control in (\ref{equ:gcc_only_gravity})\cite{Klu2018case}. The main reason for the degenerated performance is that disturbance torques from the unmodeled dynamics (such as the disturbance from electric cables in \cite{Klu2018case}) is considerably large compared to the gravity. 

\begin{figure}[!tbp]
\centering
\includegraphics[width=3.2in]{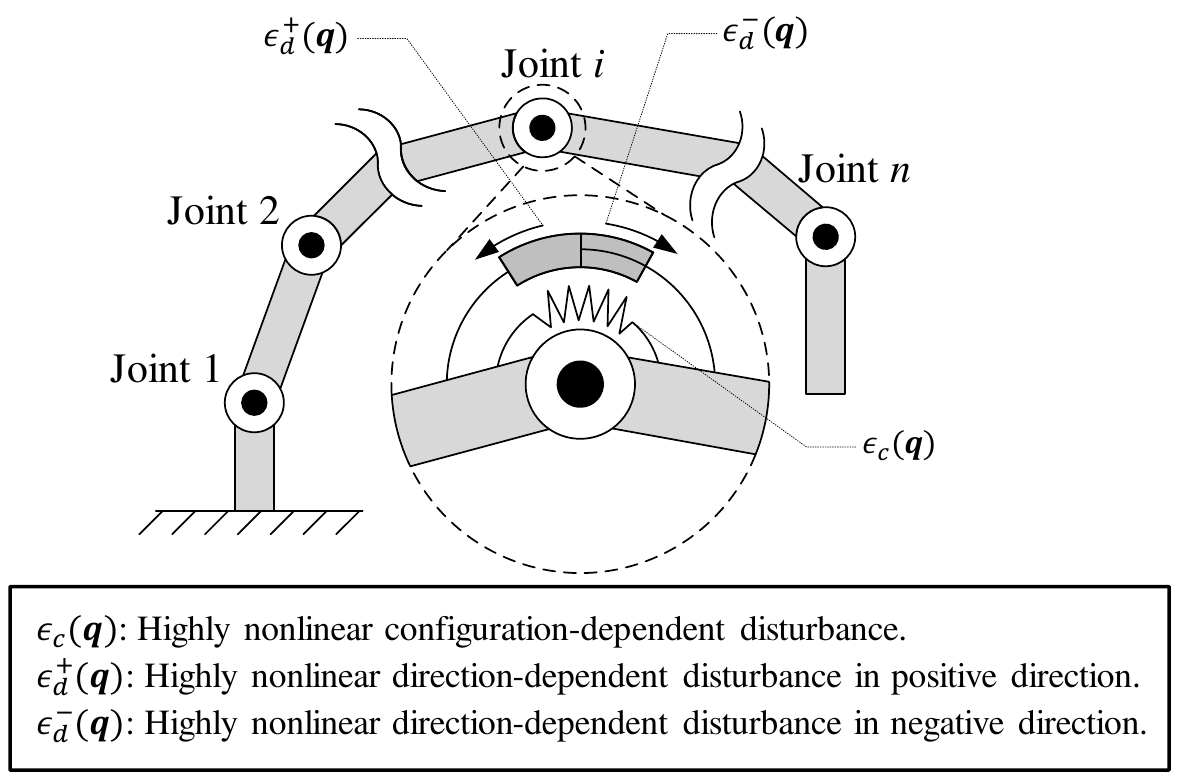}
\caption{Illustration of configuration-dependent and directional-dependent disturbance for each joint of a serial manipulator.}
\label{fig:robot_with_dist}
\end{figure}

To improve the naive control in (\ref{equ:gcc_only_gravity}), compensated torques for disturbance were taken into account in GCC. In this paper, we study the configuration-dependent direction-dependent disturbance $\boldsymbol{\epsilon(\boldsymbol{q}, \boldsymbol{\Delta q})}$ when the robot system is static (\textit{i.e.} both the velocity and acceleration of all joints are zeros). We decompose the disturbance into three parts: configuration-dependent disturbance $\boldsymbol{\epsilon_c}$, direction-dependent disturbance \textit{w.r.t.} positive joint direction $\boldsymbol{\epsilon_d^{+}}$ and that \textit{w.r.t.} negative joint direction $\boldsymbol{\epsilon_d^{-}}$. $\boldsymbol{\epsilon}$ can be formulated as
\begin{equation}
\label{equ:disturbance_formulate}
    \boldsymbol{\epsilon} =  \boldsymbol{\epsilon_c}(\boldsymbol{q}) + \boldsymbol{\epsilon_d^{+}}(\boldsymbol{q})\odot u(\Delta \boldsymbol{q}) + \boldsymbol{\epsilon_d^{-}}(\boldsymbol{q}) \odot (1-u(\Delta \boldsymbol{q})),
\end{equation}
where $\odot$ denotes Hadamard product, joint direction $\Delta \boldsymbol{q}$ is the difference between current joint position and joint position in the last iteration of the control loop, $u(\cdot)$ is the step function to determine the directions. Fig. \ref{fig:robot_with_dist} shows the illustrate three types of disturbances for each joint of a serial manipulator. In the figure, the configuration-dependent disturbance is equal to the force from highly nonlinear springs, while the direction-dependent disturbance is equal to the force from highly nonlinear dampers related to joint directions instead of joint velocity. (\ref{equ:disturbance_formulate}) can be simplified as
\begin{equation}
\label{equ:disturbance_formulate_simplified}
\begin{aligned}
    \boldsymbol{\epsilon} = &\boldsymbol{\epsilon_c}(\boldsymbol{q}) (u(\Delta \boldsymbol{q}) + 1 - u(\Delta \boldsymbol{q}))+\\
    & \boldsymbol{\epsilon_d^{+}}(\boldsymbol{q})\odot u(\Delta \boldsymbol{q}) + \boldsymbol{\epsilon_d^{-}}(\boldsymbol{q}) \odot (1-u(\Delta \boldsymbol{q}))\\
    = &  (\boldsymbol{\epsilon_c}(\boldsymbol{q}) + \boldsymbol{\epsilon_d^{+}}(\boldsymbol{q}))\odot u(\Delta \boldsymbol{q}) + \\ & (\boldsymbol{\epsilon_c}(\boldsymbol{q})+\boldsymbol{\epsilon_d^{-}}(\boldsymbol{q})) \odot (1-u(\Delta \boldsymbol{q}))\\
    =&\boldsymbol{\epsilon^{+}}(\boldsymbol{q})\odot u(\Delta \boldsymbol{q}) + \boldsymbol{\epsilon^{-}}(\boldsymbol{q}) \odot (1-u(\Delta \boldsymbol{q})),
\end{aligned}
\end{equation}
where $\boldsymbol{\epsilon^{+}}=\boldsymbol{\epsilon_c}+\boldsymbol{\epsilon_d^{+}}$ and $\boldsymbol{\epsilon^{-}}=\boldsymbol{\epsilon_c}+\boldsymbol{\epsilon_d^{-}}$ are disturbances \textit{w.r.t.} positive direction $\boldsymbol{\epsilon^{+}}$ and disturbances \textit{w.r.t.} negative direction $\boldsymbol{\epsilon^{-}}$, respectively. Additional torques to compensate disturbance in (\ref{equ:disturbance_formulate_simplified}) are added to (\ref{equ:gcc_only_gravity}) for improving the naive control as
\begin{equation}
\label{equ:GCC_with_dist}
\begin{aligned}
    \boldsymbol{\tau_c} =& \boldsymbol{g}(\boldsymbol{q}) + \boldsymbol{\epsilon}(\boldsymbol{q},\Delta\boldsymbol{q})\\
    =&  \boldsymbol{g}(\boldsymbol{q}) + \boldsymbol{\epsilon}^{+}(\boldsymbol{q})\odot u(\Delta \boldsymbol{q}) + \boldsymbol{\epsilon}^{-}(\boldsymbol{q}) \odot (1-u(\Delta \boldsymbol{q}))
\end{aligned}
\end{equation}
To achieve the control in (\ref{equ:GCC_with_dist}), researchers proposed to estimate gravity using a physics-based model and to approximate disturbance using either linear models\cite{fontanellivinci} or polynomial models\cite{wang2019dynamic,lin2019reliable}. In our previous work, we showed that polynomial models were able to provide a higher model capacity to capture the nonlinearities of disturbance compared to linear models\cite{lin2019reliable}. However, \cite{fontanellivinci,wang2019dynamic,lin2019reliable} were limited by the assumption that the configuration-dependent disturbance for each joint was only related to its own joint position. In this paper, we solve the limitation by using FDNN for a general configuration-dependent disturbance which is related to all joint positions.

The rest of the paper is organized as follows. We first introduce our method to model the gravitational dynamics with disturbance using FDNN in Section \ref{sec:modeling}, followed by learning methods in Section \ref{sec:learning}. The controller for GCC is designed in Section \ref{sec:controller}. The efficiency of our method is validated by extensive experiments of simulation and real hardware in Section \ref{sec:experiment}. Discussion and conclusion are given in Section \ref{sec:discussion} and \ref{sec:conclusion}, respectively.

\section{Modeling Gravitational Dynamics with Disturbance using FDNNs}
\label{sec:modeling}
The compensated torque in (\ref{equ:GCC_with_dist}) can be rewritten as
\begin{equation}
\label{equ:GCC_with_compensated_torques}
\begin{aligned}
    \boldsymbol{\tau_c} 
    =& \boldsymbol{g}(\boldsymbol{q})(1-u(\Delta \boldsymbol{q})+u(\Delta \boldsymbol{q})) +\\ &\boldsymbol{\epsilon}^{+}(\boldsymbol{q})\odot u(\Delta \boldsymbol{q}) + \boldsymbol{\epsilon}^{-}(\boldsymbol{q}) \odot (1-u(\Delta \boldsymbol{q}))\\
    =&(\boldsymbol{g}(\boldsymbol{q}) + \boldsymbol{\epsilon}^{+}(\boldsymbol{q}))\odot u(\Delta \boldsymbol{q}) + \\
    &(\boldsymbol{g}(\boldsymbol{q}) + \boldsymbol{\epsilon}^{-}(\boldsymbol{q})) \odot (1-u(\Delta \boldsymbol{q}))\\
    =&\boldsymbol{\tau_c}^{+}(\boldsymbol{q})\odot u(\Delta \boldsymbol{q})+ \boldsymbol{\tau_c}^{-}(\boldsymbol{q}) \odot (1-u(\Delta \boldsymbol{q})),
\end{aligned}
\end{equation}
where $\boldsymbol{\tau_c}^{+} = \boldsymbol{g} + \boldsymbol{\epsilon^{+}}$ and $\boldsymbol{\tau_c}^{-} =\boldsymbol{g}+\boldsymbol{\epsilon^{-}}$ are compensated torques \textit{w.r.t.} positive and negative directions, respectively. $\boldsymbol{\tau_c^+}$ and $\boldsymbol{\tau_c^-}$ are approximated by two sets of FDNNs as  
\begin{equation}
\label{equ:modeling dynmaics}
    \boldsymbol{\hat{\tau}_c} =\mathcal{G}(\boldsymbol{q}; \boldsymbol{W^{+}})\odot u(\Delta \boldsymbol{q})+\mathcal{G}(\boldsymbol{q}; \boldsymbol{W^{-}})\odot (1-u(\Delta \boldsymbol{q})),
\end{equation}
where $\hat{\cdot}$ refers to an approximation, $\mathcal{G}$ denotes continuous functional mapping of an FDNN, $\boldsymbol{W^{+}}$ and $\boldsymbol{W^{-}}$ are matrices of weight parameters for FDNNs $\textit{w.r.t.}$ $\boldsymbol{\hat{\tau_c}^{+}}$ and $\boldsymbol{\hat{\tau_c}^{-}}$, respectively.

To achieve stable and efficient control using FDNNs, input-output signals of FDNNs require data pre-processing through a sequence of transformations. We will discuss the transformations in the following sub-sections.

\subsection{Trigonometric Representation for Joint Position}
The $\boldsymbol{q}$ in (\ref{equ:modeling dynmaics}) is traditionally represented using a minimum representation as $\boldsymbol{q} = \begin{bmatrix} q_1 & q_2 &\dots &q_n \end{bmatrix}^{T}\in \mathbb{R}^n,
$ where the position of the $i^\mathrm{th}$ joint is $q_i=\theta_i\in [0, 2\pi)$ for a revolute joint and $q_i=d_i\in \mathbb{R}$ for a prismatic joint without considering joint limits. However, the minimal representation cannot guarantee the continuity of the functional mapping $\mathcal{G}$ in joint space due to the singularities of the representation(\textit{i.e} the neighborhood of $0$ for $q_i$ in the case of a revolute joint)\cite{lynch2017modern}, which might lead to unstable performance (Prove can be found in \cite{supplement}). To overcome the limitation, we use a non-minimum representation, trigonometric representation, where $\boldsymbol{q}$ is transformed as
\begin{equation}
\label{equ:trigon_z}
\begin{aligned}
  \boldsymbol{z} =& \Phi(\boldsymbol{q})
  = [\Phi(q_1), \Phi(q_2), \dots ,\Phi(q_n)]^T \in \mathbb{R}^{n+k},  
\end{aligned}
\end{equation}
where $k$ is the number of revolute joints for a robot; The function for the trigonometric transformation $\Phi$ is defined as 
\begin{equation}
    \Phi(x) = \begin{cases} [\sin(x), \cos(x)]^T & \text{for a revolute joint}\\
    x & \text{for a prismatic joint}
    \end{cases};
\end{equation}
Compared to the minimal representation, the trigonometric representation can guarantee the continuity of $\mathcal{G}$ in joint space, which was also proved in \cite{supplement}. 

\subsection{Input-Output Normalization}
Although an FDNN can generate a mapping between raw input and output with controlled accuracy, in practice the performance of an FDNN can be improved by normalization to reduce the risk of getting stuck in poor local minima\cite{bishop1995neural}. Furthermore, normalization contributes to preventing overfitting for joints with larger outputs in the system.

Zero-score Normalization is utilized to scale signals into a Gaussian distribution with zero mean and unit standard deviation. The zero-score function is defined as
\begin{equation}
\label{equ:forwardNormalized}
 \boldsymbol{{}^{n}a} = f_n(\boldsymbol{a}; \boldsymbol{\mu}, \boldsymbol{\sigma}) =(\boldsymbol{a}-\boldsymbol{\mu})\oslash\boldsymbol{\sigma},
\end{equation}
where $\oslash$ is Hadamard division, $\boldsymbol{\mu}$ and $\boldsymbol{\sigma}$ denote the mean and the standard deviation, respectively, $\boldsymbol{{}^{n}}(\cdot)$ denotes a normalized value and $\boldsymbol{a}$ is an input vector. Similarly, the inverse of the zero-score function is given as
\begin{equation}
\label{equ:inverseNormalized}
\boldsymbol{a} = f_n^{-1}(\boldsymbol{{}^{n}a}; \boldsymbol{\mu}, \boldsymbol{\sigma}) 
    = \boldsymbol{{}^{n}a}+\boldsymbol{\mu}\odot\boldsymbol{\sigma}.
\end{equation}

The signals in (\ref{equ:modeling dynmaics}) and (\ref{equ:trigon_z}) can be normalized as
\begin{equation}
\label{equ:normalizedInput}
    \boldsymbol{{}^{n}z} = f_n(\boldsymbol{z};\boldsymbol{\mu_{in}}, \boldsymbol{\sigma_{in}}),
\end{equation}
\begin{equation}
\label{equ:normalizedOutput}
     \boldsymbol{{}^n\hat{\tau_{c}}}=
     f_n(\boldsymbol{\hat{\tau_{c}}};\boldsymbol{\mu_{out}},\boldsymbol{\sigma_{out}}),
\end{equation}
where $\boldsymbol{\mu_{in}}$ and $\boldsymbol{\sigma_{in}}$ are parameter vectors of mean and standard deviation for the input signals $\boldsymbol{z}$, while $\boldsymbol{\mu_{out}}$ and $\boldsymbol{\sigma_{out}}$ are parameter vectors of mean and standard deviation for the output signals $\boldsymbol{\tau_c}$. Instead of mapping raw input-output signals in (\ref{equ:modeling dynmaics}), we map the normalized input-output signals by two sets of FDNNs as
\begin{equation}
\label{equ:normalizedDynamicMapping}
\begin{aligned}
\boldsymbol{{}^n\hat{\tau_{c}}} =&\boldsymbol{{}^{n}\hat{\tau}_c}^{+}(\boldsymbol{{}^n z})\odot u(\Delta \boldsymbol{q}) + \boldsymbol{{}^{n}\hat{\tau}_c}^{-}(\boldsymbol{{}^n z}) \odot (1-u(\Delta \boldsymbol{q}))
\\ =&\mathcal{G}(\boldsymbol{{}^n z}; \boldsymbol{{}^{n}W_z^{+}})\odot u(\Delta \boldsymbol{q})+\\
&\mathcal{G}(\boldsymbol{{}^n z}; \boldsymbol{{}^{n}W_z^{-}})\odot (1-u(\Delta \boldsymbol{q})),
\end{aligned}
\end{equation}
where $\boldsymbol{{}^{n}W_z^{+}}$ and $\boldsymbol{{}^{n}W_z^{-}}$ are matrices of parameters of the FDNNs for the normalized mapping corresponding to positive and negative directions, respectively. Combining (\ref{equ:normalizedInput}), (\ref{equ:normalizedDynamicMapping}) and the inverse of (\ref{equ:normalizedOutput}), the estimated compensated torques in (\ref{equ:modeling dynmaics}) can be rewritten as
\begin{equation}
\begin{aligned}
\label{equ:equ:modeling dynmaics with trigonometric and normalization}
     \boldsymbol{\hat{\tau_{c}}} =
     f_n^{-1}(&\mathcal{G}(f_n(\boldsymbol{z};\boldsymbol{\mu_{in}},\boldsymbol{\sigma_{in}}); \boldsymbol{{}^{n}W_z^{+}})\odot u(\Delta \boldsymbol{q})+\\
     &\mathcal{G}(f_n(\boldsymbol{z};\boldsymbol{\mu_{in}},\boldsymbol{\sigma_{in}}); \boldsymbol{{}^nW_z^{-}}) \odot (1-u(\Delta \boldsymbol{q}));\\
     &\boldsymbol{\mu_{out}},\boldsymbol{\sigma_{out}}).
\end{aligned}
\end{equation}

$\boldsymbol{{}^{n}W_z^{+}}$, $\boldsymbol{{}^{n}W_z^{-}}$, $\boldsymbol{\mu_{in}}$, $\boldsymbol{\sigma_{in}}$, $\boldsymbol{\mu_{out}}$ are $\boldsymbol{\sigma_{out}}$ are parameters of the predicted model for GCC. We will discuss how to learn the parameters in Section \ref{sec:learning}. 

\section{Learning FDNNs for Dynamics Modeling}
\label{sec:learning}
In this section, we will elaborate on two forms of learning schemes, Learning from Scratch (LfS) and learning with Physical Knowledge Distillation (PKD). 
\subsection{Learning from Scratch (LfS)}
To learn gravity and the disturbance of a robot system, observed data of both joint positions and joint torques should be sampled from the robot system. The data set for the observed data is defined as
\begin{equation}
\label{equ:dataset_system}
\begin{aligned}
    \mathcal{D}_{system} = \lbrace \boldsymbol{x_1^i} \gets \Phi(\boldsymbol{{q}^i}),  \boldsymbol{x_2^i} \gets \boldsymbol{{\Delta q}^{i}},
    \boldsymbol{y^i} \gets \boldsymbol{{\tau_c}^i} \rbrace_{i=1}^{T^s},
\end{aligned}
\end{equation}
where $\boldsymbol{{q}^i}$, $\boldsymbol{\Delta{q}^i}$ and $\boldsymbol{{\tau_c}^i}$ represent $\boldsymbol{q}$, $\boldsymbol{\Delta q}$ and $\boldsymbol{{\tau_c}}$ at the $i^{\mathrm{th}}$ configuration, respectively, $\boldsymbol{x_1^i}$ and $\boldsymbol{x_2^i}$ are the $i^{\mathrm{th}}$ pair of sampled input vectors while $\boldsymbol{y^i}$ is the $i^{\mathrm{th}}$ sampled output vector, $T^s$ is the total amount of sampled data in $\mathcal{D}_{system}$. 

The parameters of the input-output normalization are identified by properties of the distribution of $(\boldsymbol{x_1}, \boldsymbol{y})\in  \mathcal{D}_{system}$, where $\boldsymbol{\mu_{in}}$ and $\boldsymbol{\sigma_{in}}$ are obtained by the mean and standard deviation of $\boldsymbol{x_1}$ in $\mathcal{D}_{system}$, while $\boldsymbol{\mu_{out}}$ and $\boldsymbol{\sigma_{out}}$ are obtained by the mean and standard deviation of $\boldsymbol{y}$ in $\mathcal{D}_{system}$. Here, we do not normalize the $\boldsymbol{\Delta q}$ signal since it is not included in the input or output of FDNN in (\ref{equ:equ:modeling dynmaics with trigonometric and normalization}). Therefore, the normalized data set of $\mathcal{D}_{system}$ can be obtained by
\begin{equation}
\label{equ:DataSampleWithNorm}
\begin{aligned}
    {}^n\mathcal{D}_{system} = \lbrace \boldsymbol{x_1} &\gets (f_n(\boldsymbol{x_1}; \boldsymbol{\mu_{in}},\boldsymbol{\sigma_{in}}), ~
    \boldsymbol{x_2} \gets \boldsymbol{x_2},\\
    \boldsymbol{y} &\gets f_n(\boldsymbol{y}; \boldsymbol{\mu_{out}},\boldsymbol{\sigma_{out}}) \rbrace _{(\boldsymbol{x_1}, \boldsymbol{x_2}, \boldsymbol{y}) \in \mathcal{D}_{system}}.
\end{aligned}
\end{equation}

$\boldsymbol{{}^{n}W_z^{+}}$ and $\boldsymbol{{}^{n}W_z^{-}}$, are learned by minimizing an loss function of the LfS, which is defined as
\begin{equation}
\label{equ:lossofsystem}
    \mathcal{L}^{s} = \frac{1}{T^s}\sum || \boldsymbol{{}^n\hat{\tau_{c}}}(\boldsymbol{x}; \boldsymbol{{}^{n}W_z^{+}}, \boldsymbol{{}^{n}W_z^{-}}) - \boldsymbol{y}  ||_{(\boldsymbol{x},\boldsymbol{y}) \in {}^n\mathcal{D}_{system}},
\end{equation}
where $\boldsymbol{x}$ is the combination of $\boldsymbol{x_1}$ and $\boldsymbol{x_2}$. The loss function with regularization is given as
\begin{equation}
    \mathcal{L}^{s}_{r} = \mathcal{L}^s+\mathcal{R}(\boldsymbol{{}^{n}W_z^{+}},\boldsymbol{{}^{n}W_z^{-}}),
\end{equation}
where $\mathcal{R}$ is a regularization function. In our experiment, we used $L^2$ Norm as the regularization function.

\subsection{Learning with Physical Knowledge Distillation (PKD)}

Since the LfS approach is limited by the problem of sampling complexity for high-DOF robots, we propose a high data-efficient method using KD to learn FDNN effectively. KD was originally proposed to solve the model compression problem for classification\cite{hinton2015distilling}. In particular, the objective function was designed as the weighted average of the cross entropy of Student Model with the soft target of a TM (obtained by scaling the output of the TM using Temperature Scaling) and that with labeled correct data. In this paper, Temperature Scaling is not included in our method since it showed no improvement for regression as discussed in\cite{saputra2019distilling}. 

Compared with the nerual-network-based TM in the origin version, we use the state-of-art physics-based model in\cite{lin2019reliable} as TM, called Physical Teacher Model (PTM). (\ref{equ:GCC_with_compensated_torques}) was modelled by the PTM analytically, denoted as a function $\boldsymbol{{}^{p}{\tau_c}}(\cdot)$ that maps from $\boldsymbol{q}$ and $\boldsymbol{\Delta q}$ to compensated torques.

To distill the knowledge of a PTM, we randomly sampled $T^p$ data points from the PTM within its state space. For the $i^{\mathrm{th}}$ data point, given the inputs $\boldsymbol{{q}^i}$ and $\boldsymbol{{\Delta q}^{i}}$, the function of the PTM can predict the output torque for compensation, denoted as $\boldsymbol{{{}^p\tau_c}(\boldsymbol{{q}^i}, \boldsymbol{{\Delta q}^{i}})}$. We define a data set $\mathcal{D}_{PTM}$ that collects all the sampled data ($\boldsymbol{{q}}$, $\boldsymbol{{\Delta q}}$, $\boldsymbol{{{}^p\tau_c}(\boldsymbol{{q}}, \boldsymbol{{\Delta q}})}$) from the PTM using the representation in (\ref{equ:dataset_system}).

The parameters of the normalization are identified by the distribution property of the joint data set $\mathcal{D}_{joint} =\mathcal{D}_{system} \cup \mathcal{D}_{PTM}$. In particular, $\boldsymbol{\mu_{in}}$ and $\boldsymbol{\sigma_{in}}$ are obtained by the mean and standard deviation of $\boldsymbol{x_1}$ in $\mathcal{D}_{joint}$, while $\boldsymbol{\mu_{out}}$ and $\boldsymbol{\sigma_{out}}$ are obtained by the mean and standard deviation of $\boldsymbol{y}$ in $\mathcal{D}_{joint}$. Then, we normalize $\mathcal{D}_{PTM}$ using (\ref{equ:DataSampleWithNorm}) as ${}^n\mathcal{D}_{PTM}$.



The loss function of the PTM can be represented as
\begin{equation}
\label{equ:LossofPTM}
    \mathcal{L}^{p} = \frac{1}{T^p}\sum || \boldsymbol{{}^n\hat{\tau_{c}}}(\boldsymbol{x}; \boldsymbol{{}^{n}W_z^{+}}, \boldsymbol{{}^{n}W_z^{-}}) - \boldsymbol{y}  ||_{(\boldsymbol{x},\boldsymbol{y}) \in {}^n\mathcal{D}_{PTM}},
\end{equation}
Combining (\ref{equ:lossofsystem}) and (\ref{equ:LossofPTM}), the parameters of FDNN can be learned by minimizing the loss function with regularization for PKD, which can be represented as
\begin{equation}
    \mathcal{L}_{r}^{^{PKD}} =(1-\lambda)\mathcal{L}^{s}+ \lambda{}\mathcal{L}^{p}+\mathcal{R}(\boldsymbol{{}^{n}W_z^{+}},\boldsymbol{{}^{n}W_z^{-}}).
\end{equation}
where $\lambda \in  [0,1]$ is a coefficient value for PKD. A high value of $\lambda$ is related to high confidence in the PTM. 

Our learning method (Algorithm 1) can be summarized as following:
\begin{enumerate}
\item Sampling PTM: Sample $\boldsymbol{{}^{p}{\tau_c}}(\cdot)$ given $T^p$ randomized inputs ($\boldsymbol{q}$, $\boldsymbol{\Delta q}$) (line 3-4) and collect sampled data to $\mathcal{D}_{PTM}$ (line 5). 
\item Identifying normalized parameters and normalizing datasets: Create joint dataset (line 7) and calculate the parameters based on the distribution of the joint dataset (line 8-9). Normalize $\mathcal{D}_{system}$ and $\mathcal{D}_{PTM}$ based on the normalized parameters (line 10-11).
\item Optimizing the objective function for PKD (line 12).
\end{enumerate}

\begin{algorithm}[!tbp]
 \caption{Training FDNNs for GCC using PKD}
 \begin{algorithmic}[1]
 \renewcommand{\algorithmicrequire}{\textbf{Input:}}
 \renewcommand{\algorithmicensure}{\textbf{Output:}}
 \REQUIRE  $\boldsymbol{{}^{n}W_z^{+}}$, $\boldsymbol{{}^{n}W_z^{-}}$, $\boldsymbol{{}^{p}{\tau_c}}(\cdot)$, $\mathcal{D}_{system}$, $T^p$
 \ENSURE  $\boldsymbol{{}^{n}W_z^{+^{*}}}$, $\boldsymbol{{}^{n}W_z^{-^{*}}}$, $\boldsymbol{\mu_{in}}$, $\boldsymbol{\sigma_{in}}$, $\boldsymbol{\mu_{out}}$, $\boldsymbol{\sigma_{out}}$
    \STATE $\mathcal{D}_{PTM} = \varnothing$ 
    \FOR {$i = 1$ to $T^p$}
    \STATE Randomize $(\boldsymbol{{q}^i},  \boldsymbol{\Delta {q}^i})$ within the state space
    \STATE $\boldsymbol{x_1^i} \gets \Phi(\boldsymbol{{q}^i}),  \boldsymbol{x_2^i} \gets \boldsymbol{{\Delta q}^{i}},
    \boldsymbol{y^i} \gets \boldsymbol{{{}^p\tau_c}(\boldsymbol{{q}^i}, \boldsymbol{{\Delta q}^{i}})}$
    \STATE  $\mathcal{D}_{PTM} \gets \mathcal{D}_{PTM} \cup \lbrace \boldsymbol{x_1^i}, \boldsymbol{x_2^i}, \boldsymbol{y^i} \rbrace$
    \ENDFOR
    \STATE $\mathcal{D}_{joint} \gets \mathcal{D}_{system} \cup \mathcal{D}_{PTM}$
    \STATE $\boldsymbol{\mu_{in}}$, $\boldsymbol{\sigma_{in}} \gets \boldsymbol{\mu}$, $\boldsymbol{\sigma}$ of $\boldsymbol{x_1}$ in $\mathcal{D}_{joint}$
    \STATE $\boldsymbol{\mu_{out}}$, $\boldsymbol{\sigma_{out}} \gets \boldsymbol{\mu}$, $\boldsymbol{\sigma}$ of $\boldsymbol{y}$ in $\mathcal{D}_{joint}$
    \STATE ${}^n\mathcal{D}_{system} \gets$ \textit{Norm}$(\mathcal{D}_{system}, \boldsymbol{\mu_{in}}, \boldsymbol{\sigma_{in}},\boldsymbol{\mu_{out}}, \boldsymbol{\sigma_{out}}$)
    \STATE ${}^n\mathcal{D}_{PTM} \gets$ \textit{Norm}$(\mathcal{D}_{PTM}, \boldsymbol{\mu_{in}}, \boldsymbol{\sigma_{in}},\boldsymbol{\mu_{out}}, \boldsymbol{\sigma_{out}}$)
    \STATE $\boldsymbol{{}^{n}W_z^{+^{*}}},\boldsymbol{{}^{n}W_z^{-^{*}}} = \underset{\boldsymbol{{}^{n}W_z^{+}}, \boldsymbol{{}^{n}W_z^{-}}}{ \arg\min}\mathcal{L}^{PKD}_{r}$

 \RETURN $\boldsymbol{{}^{n}W_z^{+^{*}}}$, $\boldsymbol{{}^{n}W_z^{-^{*}}}$, $\boldsymbol{\mu_{in}}$, $\boldsymbol{\sigma_{in}}$, $\boldsymbol{\mu_{out}}$, $\boldsymbol{\sigma_{out}}$
 \end{algorithmic}
 \end{algorithm}
\section{Gravity Compensation Control using FDNNs}
\label{sec:controller}
\begin{figure}[!tbp]
\centering
\includegraphics[width=\linewidth]{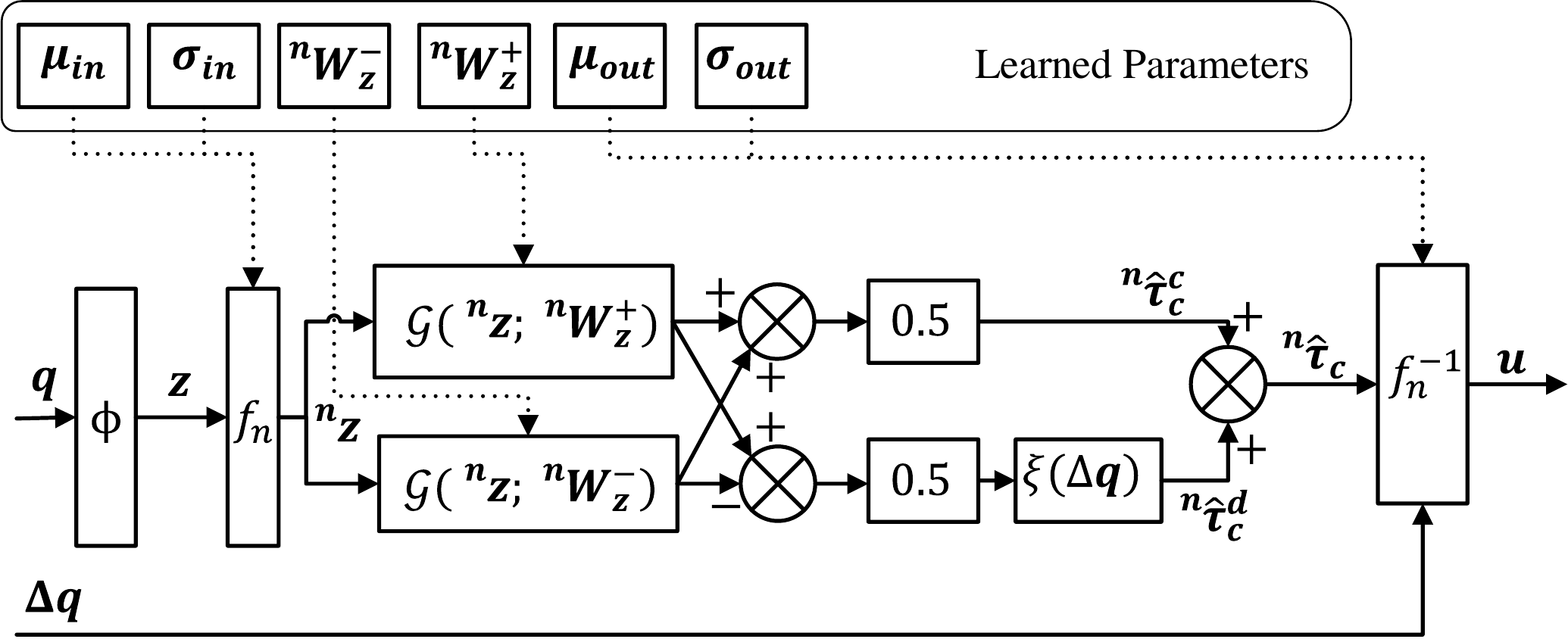}
\caption{Designed controller for GCC using FDNNs.}
\label{fig:gc_controller}
\end{figure}

We design a reliable controller for feedforward GCC using FDNNs. The designed control is similar to our previous work\cite{lin2019reliable} but in an FDNN version. The estimated configuration-dependent and direction-dependent compensated torques with normalization, $ \boldsymbol{\hat{\tau}^c_{c}}$ and $\boldsymbol{\hat{\tau}^d_{c}}$, are given as
\begin{equation}
    \boldsymbol{{}^{n}\hat{\tau}^c_{c}} = (\boldsymbol{{}^{n}\hat{\tau}_c^+}+\boldsymbol{{}^{n}\hat{\tau}_c^-})/2,
\end{equation}
\begin{equation}
    \boldsymbol{{}^{n}\hat{\tau}^d_{c}} = (\boldsymbol{{}^{n}\hat{\tau}_c^+}-\boldsymbol{{}^{n}\hat{\tau}_c^-})/2.
\end{equation}
The designed policy for GCC using FDNNs is defined as
\begin{equation}
    \boldsymbol{u} = f_n^{-1}(\boldsymbol{{}^{n}\hat{\tau}^c_c} + \boldsymbol{\xi}(\boldsymbol{\Delta q})  \odot \boldsymbol{{}^{n}\hat{\tau}^d_c};\boldsymbol{\mu_{out}}, \boldsymbol{\sigma_{out}}),
\end{equation}
where $\boldsymbol{\xi} = [\xi_1, \xi_2, \dots, \xi_n]^{T}$ is a scale vector of compensation ratio for the direction-dependent disturbance; the $i^{\mathrm{th}}$ element of $\boldsymbol{\xi}$ is designed as
\begin{equation}
\label{equ:compensated_coefficient}
    \xi_i = \begin{cases} 0 & |\Delta{q_i}|\leq \Delta{q}_{db_i} \\
     \frac{|\Delta{q}_i|-\Delta{q}_{db_i}}{\Delta{q}_{s_i}-\Delta{q}_{db_i}}\mathrm{sgn}(\Delta{q}_i)\alpha   &  \Delta{q}_{db_i} < |\Delta{q}_i| < \Delta{q}_{s_i}
     \\
    \mathrm{sgn}(\Delta{q}_i)\alpha & \Delta{q}_{s_i}  \leq  |\Delta{q}_i|
    \end{cases},
\end{equation}
where $\Delta q_i$ is the $i^{\mathrm{th}}$ element of $\boldsymbol{\Delta q}$, $\alpha$ is a compensation ratio for the directional-dependent disturbance, $\Delta{q}_{db_i}$ is a dead-band value and $\Delta{q}_{s_i}$ is a saturated value for the $i^{\mathrm{th}}$ joint. In (\ref{equ:compensated_coefficient}), when $|\Delta q_i|$ is small, $\xi_i$ is designed to be zero to reject the noise from the joint measurement. When $|\Delta q_i|$ is large, only a certain ratio $\alpha$ of the estimated directional-dependent torque is compensated. When $|\Delta q_i|$ is between $\Delta{q}_{db_i}$ and $\Delta{q}_{s_i}$, a linear interpolation is ultilized to avoid discontinuity. The design for our GCC is shown in Fig. \ref{fig:gc_controller}.

\section{Experiments}
\label{sec:experiment}
\begin{figure}[!tbp]
\centering
\includegraphics[width=0.6\columnwidth]{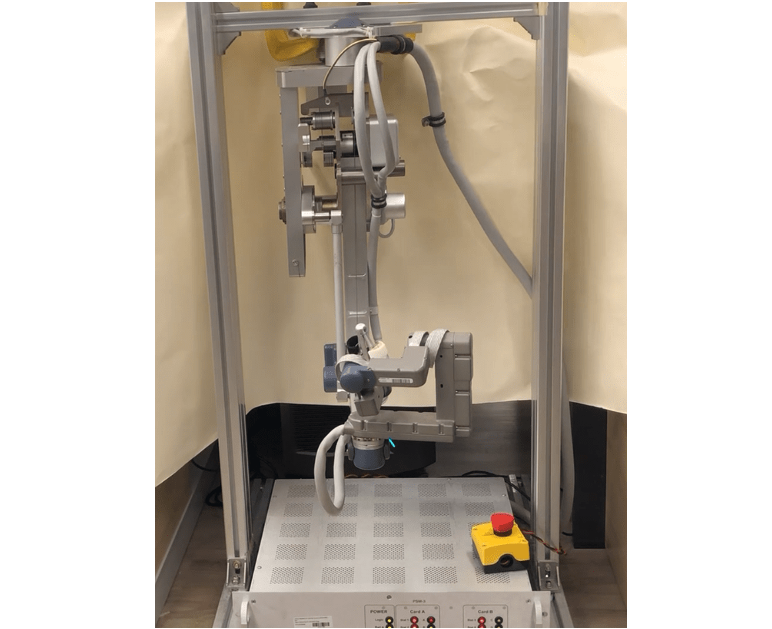}
\caption{Physical MTM setup.}
\label{fig:MTM_setup}
\end{figure}

We designed experiments to validate the effectiveness of our method based on our previous work\cite{lin2019reliable}. The evaluation was carried out on test benches consisting of Master Tool Manipulators (MTMs) of a da Vinci Research Kit (dVRK)\cite{kazanzides2014open}. Fig. \ref{fig:MTM_setup} shows our test bench. We first elaborate on the experiment settings, followed by showing the results of the experiments in both simulation and real hardware.

\subsection{Experiment Setting}
The experiments were conducted on MTMs, serial manipulators with seven motorized revolute joints and one passive pinching joint at the end. Gravity for the last two joints can be neglected due to the symmetrical structure of its link about its joint axis. In addition, the disturbance is trivial for the last two joints such that the compensated torques for the last two joints were not considered in the experiments. Thus, the MTM can be simplified as a 6-DOF serial manipulator with the mass of Link 6 lumped with the last 2 links. 

\subsubsection{Baselines}
In the experiments, we compared controllers of GCC derived from 3 types of models: the physics-based model, FDNNs trained by LfS and PKD. The analytical model served as a PTM to train the FDNNs of PKD. In practice, both Lfs and PKD can train all types of approximators for regression.  In our experiment, we used a standard architecture of FDNN: 4 fully connected layers (6,30,30,30,6) with activation function (ReLU\cite{nair2010rectified}) after each layer except the last. The number of neurons in the hidden layers of the architecture was tuned according to the learned performance empirically.

\subsubsection{Sampling Data}
Three data sets were sampled in each experiment for training, validating and testing. To sample evenly within the joint space for training data, systematic sampling (\textit{i.e.}, sampling the joint space with fixed intervals) and random sampling (\textit{i.e.}, sampling the joint space randomly) can be utilized, given the number of sampling data is large. However, when the number of sampling is small, the data distribution of the random sampling can hardly achieve the uniform sparsity over the joint space. Therefore, in the simulation where data can be infinitely sampled, we used the random sampling. In our experiments of real robots where the sampling data was expensive and limited, the systematic sampling was applied for the training data, while the random sampling was applied for the validating and testing data.

\subsubsection{Training FDNNs}
For the learning process of FDNNs, Adam \cite{kingma2014adam}, a gradient-descent optimizer, was utilized for learning the parameters of FDNNs. Furthermore, to prevent overfitting, $L_2$ Norm was utilized as the regularized function $\mathcal{R}$ and Early Stopping \cite{goodfellow2016deep} was applied based on the data set for validation. When the number of sampling data is large, these two kinds of sampling methods have similar sampling effects. For the implementation, we kept shared hyper-parameters of baselines using LfS and PKD the same to get a fair comparison.  

For the method of PKD, $T^{p}$ was set to $30000$ for dense sampling from the PTM. $\lambda$ decreased linearly during the iterative steps of the optimization. In such a manner, the guidance of PTM was dominant at the beginning of the training process, while the guidance of sampled training data increased during the process of the optimization. 
\begin{figure}[!tbp]
    \centering
        \quad
            \subfloat[\label{fig:sim_lowbias_relative_RMSE}RRMSE for simulation, low-bias PTM ]{\includegraphics[width=0.74\columnwidth]{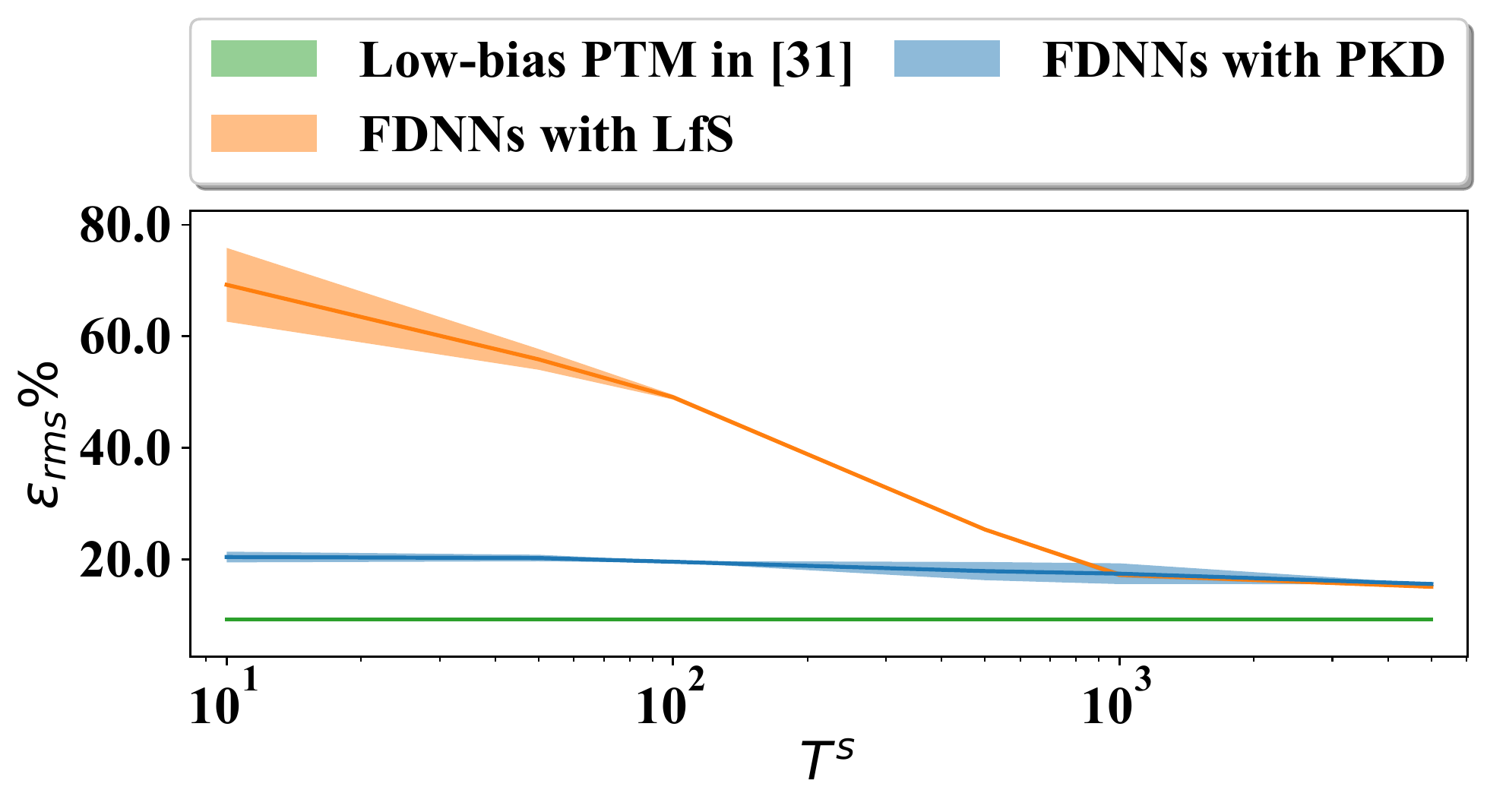}}
                \quad
            \subfloat[\label{fig:sim_highbias_Absolute_RMSE}RRMSE for simulation, high-bias PTM ]{\includegraphics[width=0.74\columnwidth]{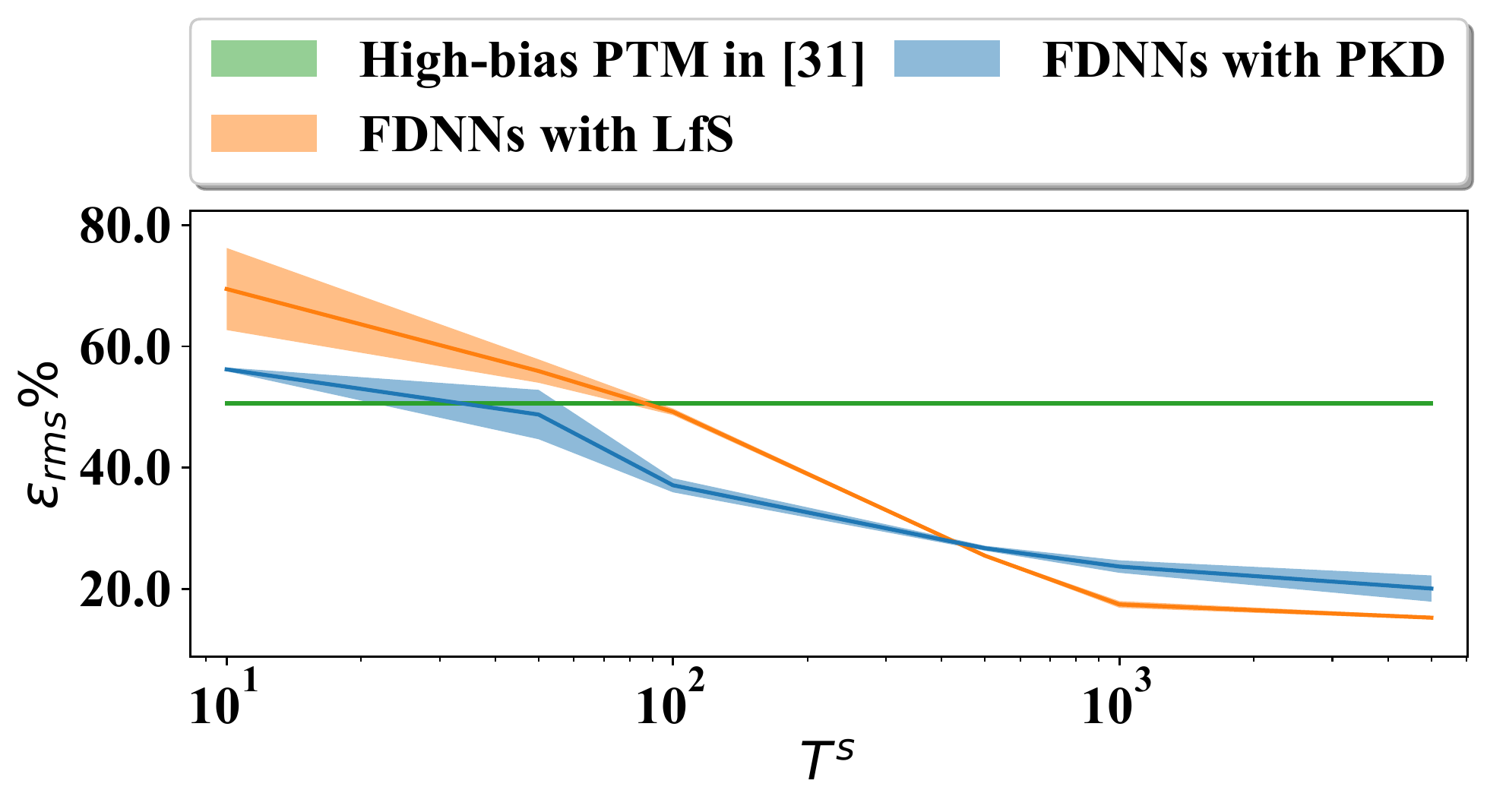}}
                
    \caption{(a-b) Average RRMSE of 6 joints \textit{w.r.t.} the amount of sampling data from the robot system using low-bias and high-bias PTMs. Lines indicate the mean of 10 repetitive experiments and shaded areas highlight the ranges within 1 standard deviation of the means. The physics-based model in \cite{lin2019reliable} with different estimation biases served as PTM. According to the results, PKD outperforms LfS in terms of smaller variances and means of RRMSE in the cases of sparse learning data, while it has similar performance compared with LfS when the learning data is rich.}
    \label{fig:simulation_experiments}
\end{figure}
\subsubsection{Offline Evaluation}
 Relative Root Mean Square Error (RRMSE) served as an indicator for evaluating the accuracy of the predicted models in GCC based on the testing data. We defined RRMSE as
\begin{equation}
    \epsilon_{RMS\%} = \sqrt{\frac{1}{N}\sum_{i=1}^{N}\frac{(\tau^i_c-\hat{\tau}^i_c)^2}{\tau_c^{i^2}}}\cdot100\%,
\end{equation}
where $\tau^i_c$ and $\hat{\tau}^i_c$ are the measured torque and the estimated torque calculated by the evaluated model for the $i^{\mathrm{th}}$ testing data, respectively; $N$ is the amount of the testing data.

\subsection{Simulation Experiments}
We designed simulation experiments to quantify the data-saving effect, where a state-of-the-art physics-based model in \cite{lin2019reliable} served as a ground-truth model to simulate gravity and disturbance of an MTM. The PTMs were obtained by adding high noises and low noises to parameters of the physics-based model for comparing cases of \enquote{Weak Teachers} (\textit{i.e.} low-bias PTMs) and \enquote{Strong Teachers} (\textit{i.e.} high-bias PTMs), respectively. In addition, noises were added to the measurement of joint positions and torques for the training and validating data. The simulated experiments explored the results in the cases of $T^s$ from $10$ to $5000$ for learning data. For each case, $10$ repetitive experiments were performed considering the randomness of the measurement noise and the learning processes.


\begin{figure}[!tbp]
    \centering
            \subfloat[\label{fig:offline_RMSE_real_MTM}RRMSE for MTM 1, comparing different baselines]{\includegraphics[width=1\columnwidth]{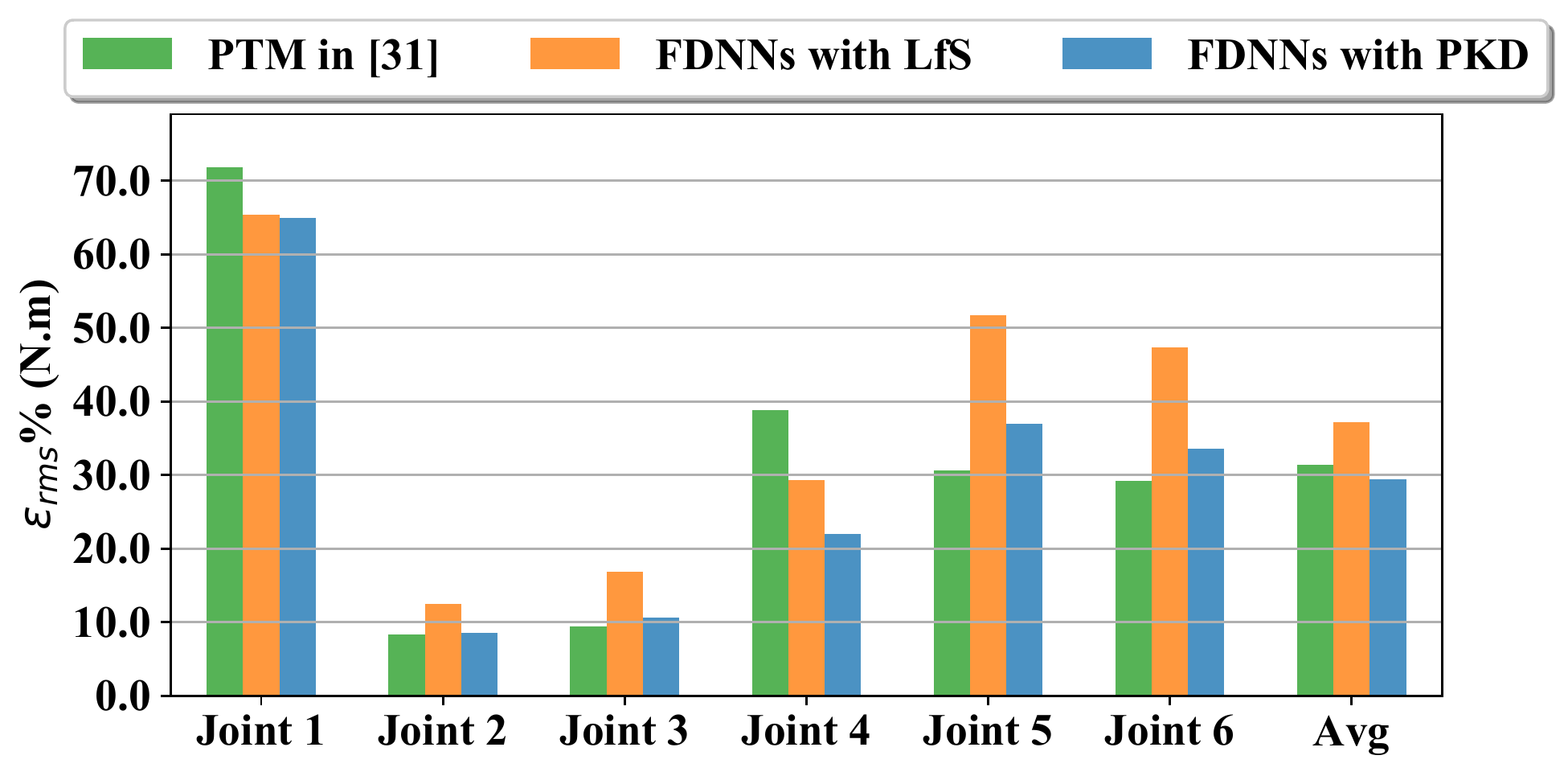}}
        \quad
            \subfloat[\label{fig:offline_Norm_Test_Real_MTM}RRMSE for MTM 1, comparing different normalized cases.]{\includegraphics[width=0.8\columnwidth]{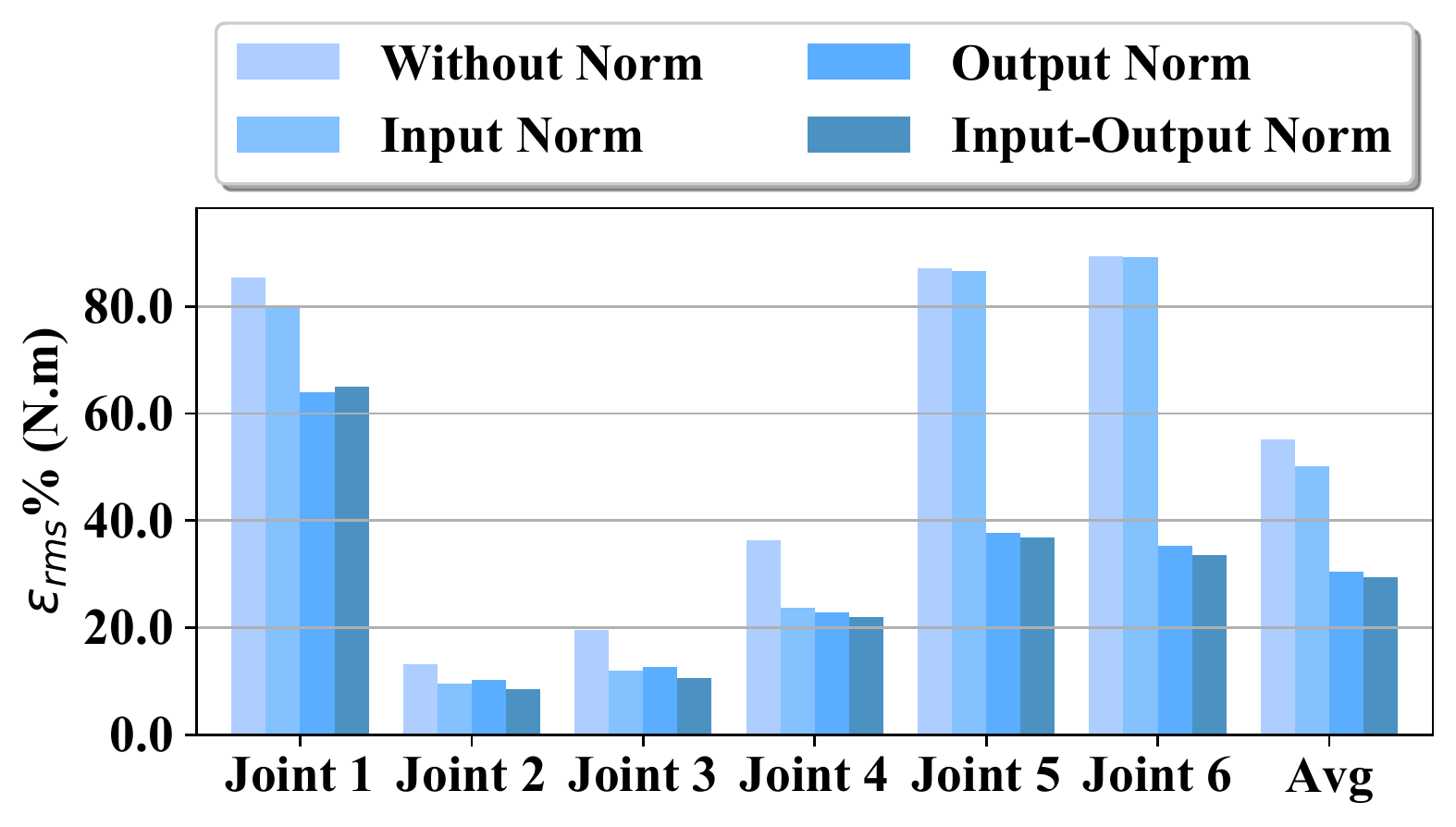}}
    \caption{Offline Test for a real MTM. (a) RRMSE for 6 joints and the average over 6 joints. (b) Average RRMSE for all joints using PKD comparing different normalized cases. In (a), PKD outperformed LfS for all joints, while it slightly outperformed the PTM in terms of the average RRMSE. In (b), both input and output normalization improved the performance.}
\end{figure}

Fig. \ref{fig:sim_lowbias_relative_RMSE} and \ref{fig:sim_highbias_Absolute_RMSE} show the results of the offline evaluation in simulation. Our approach outperformed LfS when the learning data was sparse. In the case of the high-bias PTM, the physics-based model suffered from inaccuracy of parameter estimation, leading to poor performance. Notably, our method improved the performance with increasing learning data compared to the physics-based model, while it achieved similar performance compared to the LfS approach when the learning data was rich. Our approach significantly saved the learning data compared to LfS (\textit{e.g.}, our method saved around $99\%$ training data to reach $20\%$ accuracy for average RRMSE of all joints using the low-bias PTMs).
\subsection{Real Robot Experiments}
Physical MTMs were controlled by the dVRK software system. To reach the desired state ($\boldsymbol{q}$, $\boldsymbol{\Delta q}$), we first moved the MTM to a prepared configuration $\boldsymbol{q}-\boldsymbol{\Delta q}$ and then moved to the desired position $\boldsymbol{q}$ using a PID controller. Joint positions, directions, and torques were recorded when the MTM reached a steady state. To sample the training data, we evenly sampled $4$ points for each joint for both positive and negative directions with the total amount of sampling $8192= 2\times4^6$. The total amounts of validating data and testing data were around $6000$ and $300$, respectively. We utilized the physics-based model in \cite{lin2019reliable} as the PTM.

\begin{table}[!tbp]
\newcommand{\heavy}{\cellcolor[rgb]{0,0.398,0.796}}
\newcommand{\normalColor}{\cellcolor[rgb]{0.4,0.7,1}}
\newcommand{\light}{\cellcolor[rgb]{0.797,0.896,1}}
\scriptsize
\caption{Mean and standard deviation of joint drifts averaged over 6 joints for multiple MTMs in Online Drift Test}
\label{table:online_MTMs}
\centering
\setlength{\tabcolsep}{14pt} 
\renewcommand{\arraystretch}{1.3} 
\begin{tabular}{ |c||c|c|c||}
\hline
   & PTM &    LfS &  PKD \\
\hline\hline
MTM 1       
                    &0.7  $\pm$2.4 & 6.1  $\pm$14.9 & 1.0  $\pm$5.7  \\\hline
MTM 2       
                    &2.0  $\pm$2.9 & 5.5  $\pm$12.8 & 2.1  $\pm$5.2 \\\hline
MTM 3         
                    & 1.3  $\pm$0.5 & 6.9  $\pm$6.7 & 2.0  $\pm$7.2  \\\hline

\end{tabular}
\end{table}

\begin{table}[!tbp]
\newcommand{\heavy}{\cellcolor[rgb]{0,0.398,0.796}}
\newcommand{\normalColor}{\cellcolor[rgb]{0.4,0.7,1}}
\newcommand{\light}{\cellcolor[rgb]{0.797,0.896,1}}
\scriptsize
\caption{Mean and standard deviation of Joint Drifts for MTM1 in Online Drift Test}
\label{table:online_MTM1}
\centering
\setlength{\tabcolsep}{3.5pt} 
\renewcommand{\arraystretch}{1.3} 
\begin{tabular}{ |c||c|c|c|c|c|c||}
\cline{2-7}
\hline
   & Joint 1 &  Joint 2 &  Joint 3 & Joint 4 & Joint 5 & Joint 6\\
\hline\hline
PTM      
                    &0.4  $\pm$0.8 & 0.5  $\pm$0.9 & 0.5  $\pm$0.9 & 0.6  $\pm$0.6 & 2.0  $\pm$9.7 & 0.5  $\pm$2.4\\\hline
                    LfS      
                    &0.1  $\pm$0.1 & 0.9  $\pm$2.0 & 0.9  $\pm$1.6 & 2.5  $\pm$7.0 & 24.8  $\pm$46.2 & 7.7  $\pm$14.9 \\\hline
                    PKD      
                    &0.4  $\pm$0.9 & 0.4  $\pm$0.8 & 0.4  $\pm$0.8 & 0.8  $\pm$1.4 & 2.3  $\pm$8.6 & 1.7  $\pm$5.7\\\hline
\end{tabular}
\end{table}

We designed an online test, Drift Test, based on our previous work \cite{lin2019reliable} to evaluate the performance of GCC. We tested the MTM at 400 testing points with random configurations and directions within joint limits. For each test, we first moved the MTM to a testing point using the PID controller and subsequently switched to the evaluated controller of GCC running at $500$ Hz for around 2 seconds. Both joint drifts and Cartesian drifts of the tip were recorded.

Fig. \ref{fig:offline_RMSE_real_MTM} shows the Offline Test for MTM 1. The average RRMSE of 6 joints for our method ($29.4\%$) was lower than that of other methods (PTM: $31.3\%$, LfS: $37.2\%$). In Fig. \ref{fig:offline_Norm_Test_Real_MTM}, we compared four cases to evaluate the effectiveness of the input-output normalization. Among the four cases, our approach with the input-output normalization shows the lowest RRMSE. 

Table \ref{table:online_MTMs} shows a statistical result of joint drifts averaged over 6 joints for 3 MTMs in the Online Test. Our approach reduced around 75\% the mean of the averaged drifts for 3 MTMs compared with LfS. The performance of our approach was slightly inferior to that of the state-of-art PTM in terms of the mean and standard deviation of the averaged joint drifts. However, their performance was comparable since their maximum mean of drifts for 3 MTMs were in the same scale (PTM: $2.0$; PKD: $2.1$) compared to that of LfS ($6.9$).
Table \ref{table:online_MTM1} shows the statistical result for individual joints of MTM1 in the Online Test. LfS had the largest drifts for all joints except Joint 1. The mean of drifts for Joint 5 with LfS was 10 times larger than that with PKD and PTM. On the other hand, our approach achieved similar performance with PTM or even outperformed PTM in terms of drifts of Joint 2 and Joint 3.





\section{Discussion}
\label{sec:discussion}
Our experimental results showed that PKD can significantly boost the data efficiency for FDNN in GCC. Besides GCC, our learning framework can be applied to other model-based dynamic controllers that require learning gravitational dynamics, such as PD control with gravity compensation.

While our method has successfully reduced bias of drifts of FDNN in the Online Drift Test, our variance of drifts remained large compared to that of PTM, which could be improved by ensemble learning methods. Another potential improvement is incorporating the uncertainty of PTM in our learning process.

In this work, we provide a novel and efficient method to bridge the gap between the physics-based approaches and the LfS approaches. Our method was evaluated on 6-DOF rigid-serial-link manipulators through extensive experiments, taking a key step toward exploiting our method on robot systems with higher DOFs. 




\section{Conclusion}
\label{sec:conclusion}
We proposed a novel learning framework using PKD to learn from both physical prior and observed data from the system for modeling gravitational dynamics using FDNN, where trigonometric representation and input-output normalization were applied for improving the performance of GCC. The result of experiments showed that PKD possessed better data efficiency than the standard LfS approach in GCC for high-DOF robots. In addition, it achieved competitive performance using sparse data compared to PTM.

Future directions of this work are extending our method to learning full dynamics with unknown disturbance as well as exploiting our learning framework for other types of robots (\textit{e.g.} soft robots/continuum manipulators).


\bibliographystyle{IEEEtran}

\bibliography{Reference}

\end{document}